\documentclass[letterpaper]{article} 
\usepackage{aaai2026}  
\usepackage{times}  
\usepackage{helvet}  
\usepackage{courier}  
\usepackage[hyphens]{url}  
\usepackage{graphicx} 
\usepackage{natbib}  
\usepackage{caption} 
\usepackage{algorithm}
\usepackage{algorithmic}

\usepackage{latexsym}
\usepackage{amsmath}
\usepackage{multirow}
\usepackage{booktabs}

\usepackage{newfloat}
\usepackage{listings}
\DeclareCaptionStyle{ruled}{labelfont=normalfont,labelsep=colon,strut=off} 
\floatstyle{ruled}
\newfloat{listing}{tb}{lst}{}
\floatname{listing}{Listing}
\title{X-MuTeST: A Multilingual Benchmark for Explainable Hate Speech Detection and A Novel LLM-consulted Explanation Framework}
\author{
    Mohammad Zia Ur Rehman\textsuperscript{\rm 1},
    Sai Kartheek Reddy Kasu\textsuperscript{\rm 2},
    Shashivardhan Reddy Koppula\textsuperscript{\rm 1},\\
    Sai Rithwik Reddy Chirra\textsuperscript{\rm 3},
    Shwetank Shekhar Singh\textsuperscript{\rm 4},
    Nagendra Kumar\textsuperscript{\rm 1}\thanks{Corresponding author}
}
\affiliations{
    \textsuperscript{\rm 1}Indian Institute of Technology Indore, India\\ \textsuperscript{\rm 2}Indian Institute of Information Technology Dharwad, India\\ \textsuperscript{\rm 3}Arizona State University, United States\\  \textsuperscript{\rm 4}Indian Institute of Technology Mandi, India \\
    phd2101201005@iiti.ac.in, saikartheekreddykasu@gmail.com, 
    cse210001032@iiti.ac.in,
    schirra7@asu.edu, 
    b21322@students.iitmandi.ac.in, 
    nagendra@iiti.ac.in
    
}

\usepackage{bibentry}

\begin{document}

\begin{titlepage}
\begin{center}
\vspace*{10pt}
\textbf{\Large X-MuTeST: A Multilingual Benchmark for Explainable Hate Speech Detection
and A Novel LLM-consulted Explanation Framework}
\vspace*{20pt}

Mohammad Zia Ur Rehman$^{a}$ (phd2101201005@iiti.ac.in)\\ Sai Kartheek Reddy Kasu$^b$ (saikartheekreddykasu@gmail.com)\\
Shashivardhan Reddy Koppula$^a$ (cse210001032@iiti.ac.in)\\ Sai Rithwik Reddy Chirra$^c$ (schirra7@asu.edu)\\ Shwetank Shekhar Singh$^d$ (b21322@students.iitmandi.ac.in,) \\
Nagendra Kumar$^a$ (nagendra@iiti.ac.in) \\

\hspace{1pt}

\begin{flushleft}
$^a$ Indian Institute of Technology Indore, Madhya Pradesh India\\

$^b$ Indian Institute of Information Technology Dharwad, India\\

$^c$ Arizona State University, United States\\

$^d$ Indian Institute of Technology Mandi, India\\

\vspace{2cm}
\normalsize
This is the preprint version of the accepted paper.\\
\textbf{Accepted in \textit{Proceedings of AAAI}, 2026}

\end{flushleft}        
\end{center}
\end{titlepage}

\maketitle

\begin{abstract}
Hate speech detection on social media faces challenges in both accuracy and explainability, especially for underexplored Indic languages. We propose a novel explainability-guided training framework, X-MuTeST (e\textbf{X}plainable \textbf{Mu}ltilingual ha\textbf{Te} \textbf{S}peech de\textbf{T}ection), for hate speech detection that combines high-level semantic reasoning from large language models (LLMs) with traditional attention-enhancing techniques. We extend this research to Hindi and Telugu alongside English by providing benchmark human-annotated rationales for each word to justify the assigned class label. The X-MuTeST explainability method computes the difference between the prediction probabilities of the original text and those of unigrams, bigrams, and trigrams. Final explanations are computed as the union between LLM explanations and X-MuTeST explanations. We show that leveraging human rationales during training enhances both classification performance and the model’s explainability. Moreover, combining human rationales with our explainability method to refine the model’s attention yields further improvements. We evaluate explainability using Plausibility metrics such as Token-F1 and IOU-F1 and Faithfulness metrics such as Comprehensiveness and Sufficiency. By focusing on under-resourced languages, our work advances hate speech detection across diverse linguistic contexts. Our dataset includes token-level rationale annotations for 6,004 Hindi, 4,492 Telugu, and 6,334 English samples. Data and code are available on: 
https://github.com/ziarehman30/X-MuTeST. 
\end{abstract}


\section{Introduction}
\begin{figure}[h!] 
    \centering
    \includegraphics[width=0.6\textwidth]{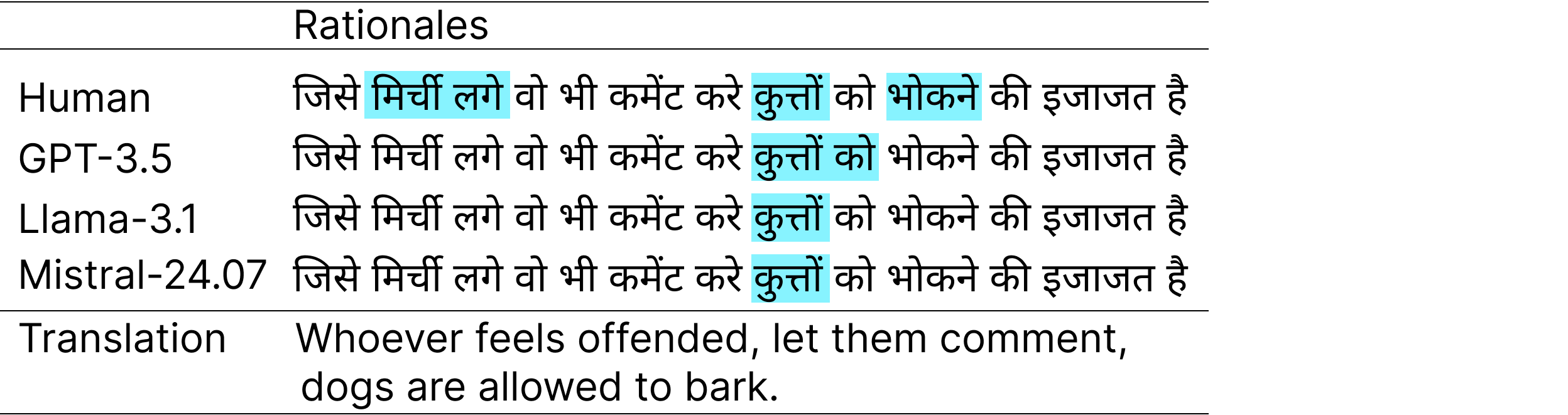} 
    \caption{Example of human and LLM rationales}
    \label{fig:example}
\end{figure}

The rise in online media usage has heightened exposure to hate speech, making detection systems increasingly essential. While initial research focused only on the detection of hateful content \cite{waseem2016hateful}, more recent efforts have shifted towards explainability and addressing the rationales behind labelling content as hateful \cite{clarke2023rule}. Advanced methods such as Large Language Models (LLMs) and their predecessors, such as BERT \cite{devlin2019bert}, can provide explanations for such decisions. LLMs can be prompted to provide the explanations in the form of an identified hateful list of words \cite{roy2023probing} for a given sentence, whereas methods such as BERT can provide explanations in the form of attention given to each word of the sentence \cite{mathew2021hatexplain}. However, in the case of under-resourced languages, where overall sequence classification performance has improved significantly over time \cite{rehman2023user}, machine-provided rationales, many a time, do not align well with human rationales. For instance, in Figure \ref{fig:example}, the translation of the first and third rationales provided by the human annotator would be ``to feel offended'' and ``to bark'', respectively, which are very offensive terms in the Hindi language. Though the LLMs correctly identify the explicit term ``dog'', they do not identify first and third rationale terms as hateful or offensive, which can be attributed to the fact that societal and cultural aspects of languages are often overlooked by these methods for under-resourced languages. This gap between machine-provided rationales and human rationales can be bridged by providing rationale resources for these languages.

\subsection{Motivation: Scarcity of Hateful Rationales}
Though numerous rationale and reasoning-based resources for the English language \cite{rajani2019explain, camburu2018snli}  are compiled in ERASER \cite{deyoung2020eraser} in various domains spanning from movie reviews \cite{zaidan2008modeling}, questions with multiple correct answers \cite{khashabi2018looking}, and to other miscellaneous domains \cite{lehman2019inferring, clark2019boolq}, there is a scarcity of rationale-based resources in the domain of hate speech detection. In 2021, \citet{mathew2021hatexplain} introduced the HateXplain dataset, which features word and phrase-level span annotations for hate speech in English, claiming to be the first of its kind. However, similar datasets for under-resourced languages such as Hindi and Telugu are nonexistent to the best of our knowledge. Even for English, HateXplain is one of the few datasets offering human rationales for hate speech detection. To address this gap, we provide word-level human rationales for hate speech detection datasets in Telugu, Hindi, and English. Our primary focus was on Indic languages. However, we believe that evaluating models on the English dataset alongside HateXplain enhances the generalizability of future approaches.

\subsection{A Glance at the Proposed Solution}
We propose a novel explainability framework that combines LLM-based and n-gram-based explanations to generate final explanations. We propose an e\textbf{X}plainable \textbf{Mu}ltilingual ha\textbf{Te} \textbf{S}peech de\textbf{T}ection ( X-MuTeST) method. X-MuTeST is a two-stage explainable hate speech detection framework designed to enhance both classification performance and explainability. It incorporates a novel explainability method based on n-grams to identify relevant tokens. In the first stage, the model's attention is guided by human-annotated rationales to align the model's focus with key tokens identified by humans. In the second stage, training is guided by the n-gram-based explainability method, which generates attention masks based on model-driven token importance. This approach balances human rationales and the model's insights, improving both explainability and classification performance. Finally, union of explanations is taken from LLama 3.1 and X-MuTeST. Through experiments on three datasets, Hindi, Telugu, and English, we demonstrate the efficacy of the proposed framework.\\
The key contributions of this work are as follows:
\begin{itemize}
    \item We contribute benchmark human-annotated rationales for under-resourced languages, such as Hindi and Telugu, alongside English. This will provide a valuable resource for future research in explainable hate speech detection across multiple languages.
    \item We propose a novel LLM-consulted and n-gram-based hybrid explainability method that combines LLMs with traditional transformer-based attention improvement. The proposed method enhances explainability performance.
    \item We propose a novel explainable hate speech detection framework with a two-stage training method that balances classification and explainability performance.
\end{itemize}

\section{Related Work}
\label{sec:related_works}

\subsection{Explainability in Hate Speech Detection}

Explainability in hate speech detection has garnered significant attention, with recent advancements highlighting both its potential and underlying challenges. The works,  such as \citet{zaidan2007using} and  \citet{yessenalina2010automatically} utilized rationales for improving sentiment classification, which set the stage for explainable approaches in more domains \cite{wang2024explainable}, including hate speech \cite{mathew2021hatexplain, lin2024towards}.
The work of \citet{mathew2021hatexplain} provides one of the first datasets to include human-annotated rationales for hate speech. For each sample, annotators not only label whether the text contains hate speech but also highlight the specific tokens that are relevant to that judgment. The authors highlight that even though a model may be accurate in classification, its reasoning may not align with human logic or provide meaningful explanations.

Recent works in LLMs and attention-based models also highlight critical challenges in explainability, as seen in works by \citet{roy2023probing},  \citet{clarke2023rule}, and \citet{arshad2024understanding}. These studies underscore difficulties in ensuring faithful rationales, even when powerful models achieve high classification accuracy. 

\subsection{Multilingual Explainable Methods}

  \citet{kapil2023hhld} and \citet{sawant2024using} emphasize the importance of precise annotations and transparency in low-resource languages such as Hindi, where cultural aspects affect how hate speech is perceived and classified. Similarly,  \citet{yadav2023hate} address the complexities of code-mixed languages, highlighting how explainability helps clarify model predictions in multilingual settings. In this context, \citet{geleta2023exploring} introduce a multilingual framework for assessing hate speech intensity levels using explainable AI, further underlining the need for transparent hate speech detection models. By leveraging explainable methodologies, these studies collectively enhance the detection of hate speech across linguistic boundaries, ensuring that cultural and linguistic differences are accounted for in model interpretation.

\section{Dataset Description and Annotation}
\label{sec:dataset}

\subsection{Dataset Source}
The Hindi and English datasets utilized in this research are derived from the Hate Speech and Offensive Content Identification (HASOC) contest, which is dedicated to the detection of hate speech and offensive language across various languages and dialects. We collect datasets from HASOC 2020 \cite{mandl2020overview, mandl2020overview1} and HASOC 2021 \cite{modha2021overview} for Hindi and English, which include samples sourced from Twitter featuring annotated instances of hate speech and offensive content. We mix datasets of two years to increase the number of samples. We select binary labels that categorize the sentences into \textit{HATE} and \textit{NOT HATE} categories. Subsequently, the testing set is split, where 15\% of the samples are retained for testing for both languages. Dataset details are shown in Table \ref{tab:dataset}.

The Telugu dataset is collected from the Hate and Offensive Language Detection in Telugu Codemixed Text (HOLD-Telugu) task \cite{priyadharshini2023overview,premjith2024findings} organized as
part of DravidianLangTech 2024 \cite{chakravarthi2024proceedings}. The comments in the dataset are collected from YouTube and annotated for binary class labels, \textit{HATE} and \textit{NON-HATE}. The original dataset contains text in Latin script; we transliterate the text into Telugu script using the IndicXlit method \cite{madhani2023aksharantar}. A testing set is provided in the dataset. Dataset details are shown in Table \ref{tab:dataset}.

\begin{table}[!ht]
\scriptsize
\centering
\resizebox{0.45\textwidth}{!}{
\begin{tabular}{lccc}
\hline
\textbf{Metric} & \textbf{Telugu} & \textbf{Hindi} & \textbf{English} \\
\hline
Total Records          & 4,492     & 6,004     & 6,334     \\
Hate Records           & 2,131     & 2,026     & 3,767     \\
Non-Hate Records       & 2,361     & 3,978     & 2,567     \\
Avg Word Count         & 6.46      & 18.84     & 19.20     \\
Avg Character Count    & 68.98     & 110.93    & 100.47    \\
Avg Rationale Length   & 3.10      & 3.26      & 3.78      \\
\hline
\end{tabular}}
\caption{Summary of hate and non-hate records}
\label{tab:dataset}
\end{table}

\subsubsection{Profile of Annotators}
To ensure the reliability and consistency of the data annotation process, we employed a rigorous selection and training procedure for our annotators. Five annotators were hired based on their linguistic proficiency, subject matter expertise, and prior experience in annotation tasks. Annotators were selected such that we had three annotators for each language. All annotators underwent a structured training phase to familiarize themselves with the annotation guidelines. Periodic quality checks were conducted, and discrepancies were resolved through consensus discussions \cite{rehman2025implihatevid}. 
Given the nature of our dataset, we carefully select annotators above the age of 18 from diverse regions across the country to mitigate potential demographic biases.

\subsection{Rationale Annotation Procedure}

\begin{figure}[h] 
    \centering
    \includegraphics[width=0.5\textwidth]{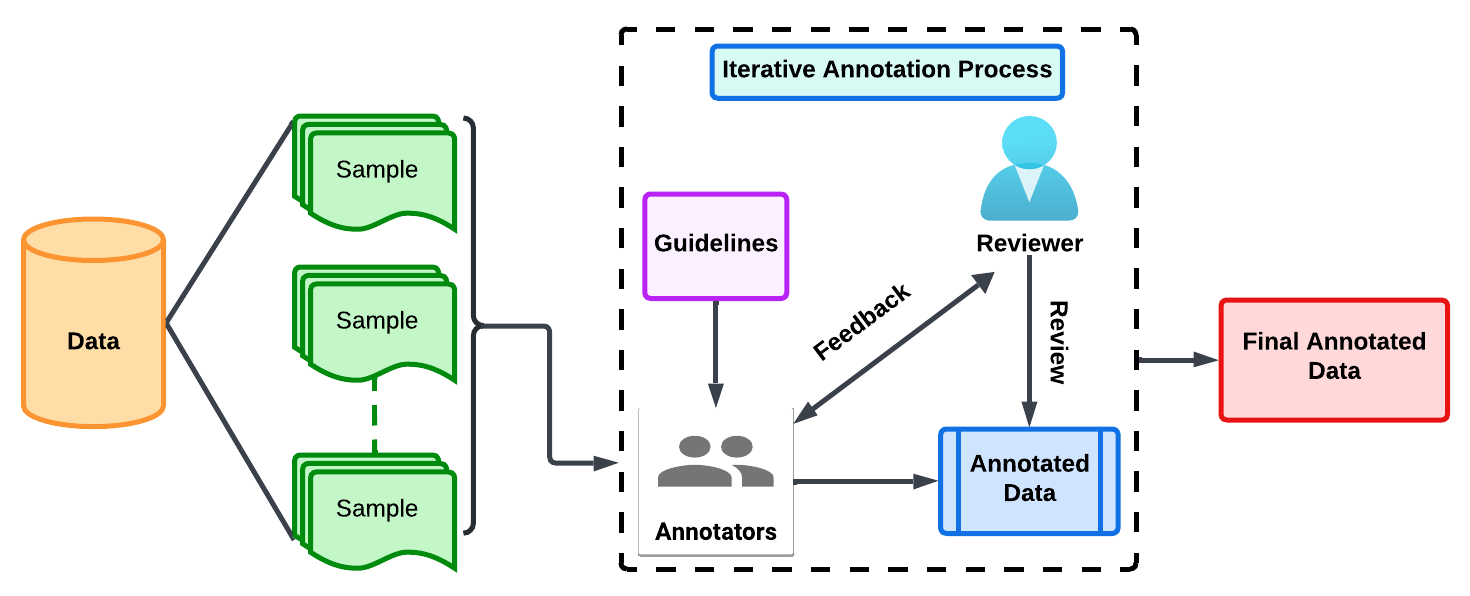} 
    \caption{Iterative annotation procedure}
    \label{fig:annotation}
\end{figure}

To ensure the quality and accuracy of annotations, we engage native language experts fluent in Hindi, English, and Telugu. This is crucial for preserving the societal and cultural aspects within the dataset. Each post is annotated by three annotators. Annotators are tasked with identifying token-level rationales that explain why a particular text is classified as hateful or non-hateful. For instance, in a sentence marked as hateful, the rationale could be an offensive word or phrase that justifies the hate speech classification. Human annotators provide these rationales by marking hate-indicating tokens with a label of 1 and non-hate tokens with 0, enhancing both model interpretability and explainability. In non-hate samples, all tokens are marked as 0.
During the initial annotation rounds, we observe \textit{moderate to substantial agreement (61-80\%).} Since our dataset is intended to support explainability in downstream models, maintaining high annotation consistency is crucial. Although annotators are initially provided with detailed guidelines, the agreement scores indicate the need for further refinement. As supported in prior works \cite{schmer2024annotator, hahn2012iterative}, iterative annotation can significantly improve label quality and annotator alignment.  To address this, we employ an iterative annotation process, as shown in Figure \ref{fig:annotation}, conducting additional guidance sessions and having annotators re-annotate the dataset, which increased the agreement among annotators. As a token of appreciation and to ensure fair compensation, annotators are provided with free access to our GPU for 150 hours, valid for one year. Each dataset is annotated by three annotators for rationales. Table \ref{tab:kappa} shows the inter-annotator agreement scores using Cohen's Kappa method between any two annotators, whereas overall agreement is computed using Fleiss method.
From Table~\ref{tab:kappa}, all three languages show substantial agreement. English exhibits the highest consistency (85.10), while Hindi shows slightly lower pairwise scores. These results confirm that iterative refinement notably improved annotation reliability across languages.

\begin{table}[!ht]
\scriptsize
\centering
\resizebox{0.48\textwidth}{!}{
\begin{tabular}{lcccc}
\hline
\textbf{Language} & \textbf{A1 \& A2} & \textbf{A2 \& A3} & \textbf{A3 \& A1} & \textbf{Overall} \\
\hline
Telugu   & 82.50  & 87.75  & \textbf{88.25} & 81.00 \\
Hindi    & 86.45 & \textbf{88.00}  & 79.20  & 83.15 \\
English  & 87.60 & 82.00  & \textbf{89.30} & 85.10 \\
\hline
\end{tabular}}
\caption{Kappa agreement scores}
\label{tab:kappa}
\end{table}

\begin{figure*}[h] 
    \centering
    \includegraphics[width=0.9\textwidth]{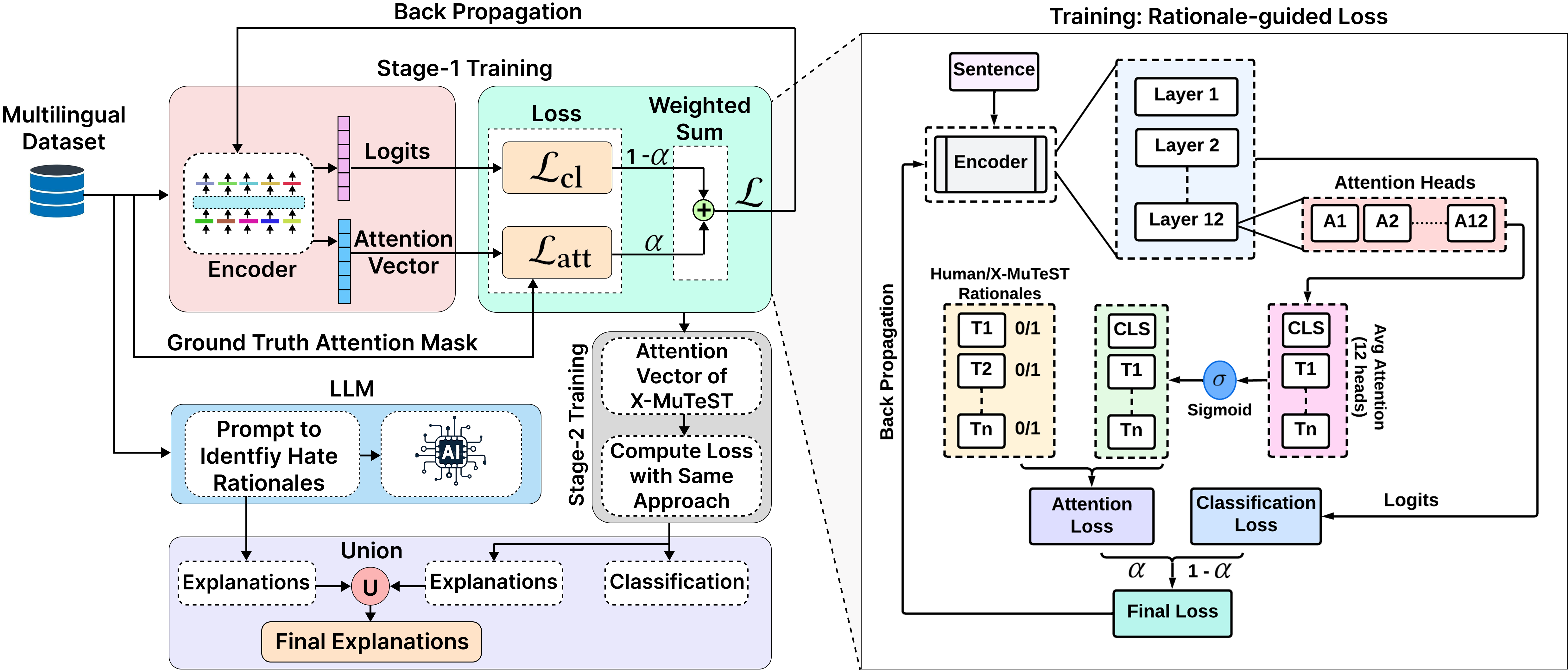} 
    \caption{Architecture of LLM-consulted explanation framework}
    \label{fig:method}
\end{figure*}

\section{Methodology}
\label{sec:methodology}

We propose \textbf{X-MuTeST}, an explainable hate speech detection framework that leverages both human rationales and a novel explainability formulation to train a model. Training proceeds in two stages as shown in Figure \ref{fig:method}. In Stage-1, the model is trained for three epochs with human rationales guiding its \textit{attention}. In Stage-2, the target attention mask is derived from our explainability method, allowing the model to balance human annotations with tokens deemed salient by the model itself. This dual approach enhances both classification accuracy and interoperability. For the final, explainability, we take the union of explanations provided by LLama-3.1 and our attention-based method. We integrate LLMs, as they have shown promise in semantic and contextual reasoning \cite{kasu2025d}. Specifically, LLaMA-3.1 is selected based on empirical evidence, as it demonstrated a higher token-level alignment with human rationales than other LLMs, for Telugu and Hindi.

For each language, we select the encoder that provides the best classification performance after finetuning. Hence, for X-MuTeST framework, we employ Muril \cite{khanuja2021muril} for Telugu and English and XLMR \cite{conneau2020unsupervised} for Hindi.

\subsection{Training Procedure}

In Stage-1, the model’s attention is guided by human rationales. Let \(a_i\) denote the normalized attention score assigned to the \(i\)th token (word) by the model and \(R_i\) the corresponding binary human rationale. The attention weights are computed using the dot product between token representations and the \([CLS]\) token, followed by softmax normalization:
\begin{equation}
a_i = \frac{\exp(h_i^\top h_{\text{[CLS]}})}{\sum_{j=1}^{L} \exp(h_j^\top h_{\text{[CLS]}})}
\end{equation}
where \(h_i\) and \(h_{\text{[CLS]}}\) are the hidden states of the \(i\)th token and the \([CLS]\) token respectively.

The attention alignment loss is computed using cross-entropy:
\begin{equation}
\mathcal{L}_{att} = -\sum_{i=1}^{L} R_i \log a_i
\end{equation}

The overall loss function combines this attention loss with the standard classification loss \(\mathcal{L}_{cl}\) as follows:
\begin{equation}
\label{eq:loss1}
\mathcal{L}_1 = \alpha\, \mathcal{L}_{att} + (1-\alpha)\, \mathcal{L}_{cl}
\end{equation}
where \(\alpha\) is a balancing coefficient (set to 0.3 in Stage-1). This helps the model to jointly learn task-specific classification while aligning its attention with human-provided rationale supervision.

In Stage-2, the human rationales are replaced by model-generated attention targets derived from our explainability method. For a sequence \(S\) of length \(L\), we identify the top \(k\) tokens based on their explainability scores, where
\begin{equation}
k = 
\begin{cases}
5, & \text{if } L \geq 10 \\
\lceil L/2 \rceil, & \text{otherwise}
\end{cases}
\end{equation}
The loss function in this stage remains the same as in Equation \eqref{eq:loss1}, but with updated \(\alpha\) values: 0.6 for Telugu and English, and 0.7 for Hindi. These values are selected through empirical tuning for best validation performance.

\subsection{Explainability via N-gram Contributions}

Our explainability method quantifies the importance of each token based on its contributions across overlapping n-grams. Let the model’s output logits for the original sequence \(S\) be:
\begin{equation}
P_{\text{orig}} = f(S)
\end{equation}
For each n-gram \(ng\) (\(n=1,2,3\)), we compute:
\begin{equation}
P_{ng} = f(ng)
\end{equation}
and define the logit drop as:
\begin{equation}
\Delta P_{ng} = \left| P_{\text{orig}} - P_{ng} \right|
\end{equation}
The unnormalized importance score for a token \(t\) is then computed as:
\begin{equation}
E[t] = \sum_{n=1}^{3} w_n \cdot \frac{1}{N_t^{(n)}} \sum_{\substack{ng \ni t}} \Delta P_{ng}
\end{equation}
where \(w_1 = 0.5\), \(w_2 = 0.3\), and \(w_3 = 0.2\) are the weights assigned to unigrams, bigrams, and trigrams respectively, and \(N_t^{(n)}\) is the number of n-grams of length \(n\) containing token \(t\). Finally, the scores are normalized:
\begin{equation}
\tilde{E}[t] = \frac{E[t]}{\sum_j E[j]}
\end{equation}
This formulation captures both token-level and contextual relevance, providing more comprehensive guidance to the model in Stage-2.

\subsection{Final Explanations through LLM Consultation}

As shown in Figure \ref{fig:method}, the final explanation is derived by combining rationales from two sources: our in-model n-gram based method and LLM-generated rationales. Let \(\mathcal{E}_{\text{X}}\) denote the set of top tokens selected by the X-MuTeST explainability scores, and \(\mathcal{E}_{\text{LLM}}\) the rationales returned by LLaMa-3.1. The final explanation set is given by:
\begin{equation}
\mathcal{E}_{\text{final}} = \mathcal{E}_{\text{X}} \cup \mathcal{E}_{\text{LLM}}
\end{equation}
To assess alignment between the two explanation sources, we also compute an agreement score:
\begin{equation}
\text{Agreement} = \frac{|\mathcal{E}_{\text{X}} \cap \mathcal{E}_{\text{LLM}}|}{|\mathcal{E}_{\text{X}} \cup \mathcal{E}_{\text{LLM}}|}
\end{equation}
This union-based approach ensures both syntactic and semantic coverage, combining task-specific saliency with LLM-derived contextual reasoning. LLaMa-3.1 is used for LLM consultation due to its superior interpretability, as discussed in the results section.

\section{Experimental Setup and Results}
\label{sec:results}

\subsection{Metrics for Evaluation}

\subsubsection{\textbf{Performance-Based Metrics}}

We use standard performance-based metrics, including accuracy, F1 score, and macro-F1, to evaluate the classification performance. 

\subsubsection{\textbf{Explainability-Based Metrics}}

We assess explainability using two metrics: \textit{plausibility} (how believable the explanation is) and \textit{faithfulness} (how well it reflects the model's decision process) \cite{jacovi2020towards, mathew2021hatexplain}.

\vspace{-1.5mm}
\paragraph{\textbf{Plausibility}}

Plausibility is measured via token-level F1-score and Intersection-over-Union (IOU) F1-score. IOU is computed as the ratio of overlapping tokens to the union of predicted and ground truth rationales, with a match defined when the overlap exceeds 0.5 \cite{deyoung2020eraser, mathew2021hatexplain}.

\vspace{-1mm}
\paragraph{\textbf{Faithfulness}}

Faithfulness is evaluated using \textit{comprehensiveness} and \textit{sufficiency} \cite{deyoung2020eraser, mathew2021hatexplain}.

\textbf{Comprehensiveness} quantifies the reliance on rationales by measuring the drop in prediction confidence when they are removed:
\begin{equation}
\text{Comprehensiveness} = P(x_i)_c - P(x_i \setminus r_i)_c
\end{equation}
where \(P(x_i)_c\) is the probability for class \(c\) using the full input \(x_i\), and \(P(x_i \setminus r_i)_c\) is the probability when the predicted rationale \(r_i\) is masked.

\textbf{Sufficiency} evaluates whether the rationale alone supports the prediction:
\begin{equation}
\text{Sufficiency} = P(x_i)_c - P(r_i)_c
\end{equation}
with \(P(r_i)_c\) computed from the rationale \(r_i\) only. Lower sufficiency indicates that the rationale is informative.

For both metrics, tokens are replaced with the \(<MASK>\) token, and the top-5 tokens are used as the predicted rationales.

\begin{table*}[!ht]
\centering
\scriptsize
\renewcommand{\arraystretch}{1.1}
\begin{tabular}{l ccc ccc cc}
\hline
\multirow{2}{*}{\textbf{Model}} & \multicolumn{3}{c}{\textbf{Performance}} & \multicolumn{3}{c}{\textbf{Plausibility}} & \multicolumn{2}{c}{\textbf{Faithfulness}} \\
\cmidrule(lr){2-4} \cmidrule(lr){5-7} \cmidrule(lr){8-9}
 & \textbf{Acc $\uparrow$} & \textbf{F1 $\uparrow$} & \textbf{Macro-F1 $\uparrow$} & \textbf{Token-F1 $\uparrow$} & \textbf{IOU-F1 $\uparrow$} & & \textbf{Comp $\uparrow$} & \textbf{Suff $\downarrow$} \\
\hline
XLMR-LIME & 0.7927 & 0.8030 & 0.7921 & 0.5143 & 0.2232 & & 0.6470 & 0.3094 \\
XLMR-XMuTeST & 0.7927 & 0.8030 & 0.7921 & 0.5256 & 0.2261 & & 0.6539 & 0.2568 \\
XLMR-Rationale-LIME & 0.8455 & 0.8355 & 0.8450 & 0.5492 & 0.2385 & & 0.7395 & 0.1380 \\
XLMR-Rationale-XMuTeST & 0.8455 & 0.8355 & 0.8450 & 0.5634 & 0.2417 & & 0.7267 & 0.1035 \\
\hline
Muril-LIME & 0.8598 & 0.8595 & 0.8591 & 0.5515 & 0.2840 & & 0.6676 & 0.1287 \\
Muril-XMuTeST & 0.8598 & 0.8595 & 0.8591 & 0.5583 & 0.2919 & & 0.6852 & 0.0947 \\
Muril-Rationale-LIME & 0.8740 & 0.8640 & 0.8733 & 0.5608 & 0.2941 & & 0.6968 & 0.0802 \\
Muril-Rationale-XMuTeST & 0.8740 & 0.8640 & 0.8733 & 0.5754 & 0.3013 & & 0.7243 & 0.0747 \\
\hline
GPT-4o & 0.6484 & 0.5929 & 0.6417 & 0.3893 & 0.1815 & & - & - \\
Llama-3.1 & 0.6362 & 0.6716 & 0.6319 & 0.5154 & 0.2806 & & - & - \\
Mistral-24.07 & 0.6402 & 0.4451 & 0.5895 & 0.3453 & 0.1228 & & - & - \\
\hline
BERT-HateXplain-LIME & 0.8089 & 0.8075 & 0.8084  & 0.5560  & 0.2233 & & 0.6851 & 0.0516 \\
BERT-ILP & 0.8213 & 0.8183 & 0.8221  & 0.5236  & 0.2440 & & 0.5833 & 0.0849 \\
\midrule
\textbf{X-MuTeST with LLM} & \textbf{0.8881} & \textbf{0.8762} & \textbf{0.8849} & \textbf{0.6231} & \textbf{0.3189} & & \textbf{0.7456} & \textbf{0.0448} \\
\bottomrule
\end{tabular}
\caption{Classification and explainability performance of various methods on Telugu}
\label{table:telugu}
\end{table*}

\subsection{Comparison Methods}
\label{sec:comparison}
We employ different multilingual transformer encoders for classification, whereas LIME and X-MuTeST methods are used for explainability. Additionally, we employ three LLMs, GPT-4o \cite{gpt}, LLaMa-3.1 \cite{dubey2024llama}, and Mistral-24.07 \cite{mistral},  using the prompt-based zero-shot method.  We use token-F1 and IOU-F1 to evaluate the explainability of LLMs. As prompt-based methods do not provide probability scores for their decision, comprehensiveness and sufficiency scores could not be computed for LLMs. We also compared our method with BERT-HateXplain-LIME~\cite{mathew2021hatexplain} and with BERT-ILP method \cite{nguyen2022towards}. We evaluate transformer encoders as follows:

  We employ two multilingual encoders Muril \cite{khanuja2021muril} and XLMR \cite{conneau2020unsupervised} for comparison. First, the base encoder model is finetuned for the sequence classification objective, and the LIME and the X-MuTeST approaches are used to evaluate explainability performance after training. 
Next, the base encoder is trained by integrating human rationales, and subsequently, LIME and the X-MuTeST method are employed for explainability.

\begin{table*}[!ht]
\centering
\scriptsize
\renewcommand{\arraystretch}{1.1}
\begin{tabular}{l ccc ccc cc}
\hline
\multirow{2}{*}{\textbf{Model}} & \multicolumn{3}{c}{\textbf{Performance}} & \multicolumn{3}{c}{\textbf{Plausibility}} & \multicolumn{2}{c}{\textbf{Faithfulness}} \\
\cmidrule(lr){2-4} \cmidrule(lr){5-7} \cmidrule(lr){8-9}
 & \textbf{Acc $\uparrow$} & \textbf{F1 $\uparrow$} & \textbf{Macro-F1 $\uparrow$} & \textbf{Token-F1 $\uparrow$} & \textbf{IOU-F1 $\uparrow$} & & \textbf{Comp $\uparrow$} & \textbf{Suff $\downarrow$} \\
\hline
XLMR-LIME     & 0.8535 & 0.7988 & 0.8418 & 0.3068 & 0.1494 & & 0.6651 & 0.7296 \\
XLMR-XMuTeST     & 0.8535 & 0.7988 & 0.8418 & 0.3948 & 0.1799 & & 0.7122 & 0.6528 \\
XLMR-Rationale-LIME    & 0.8590 & 0.8000 & 0.8456 & 0.3127 & 0.1564 & & 0.7011 & 0.5314 \\
XLMR-Rationale-XMuTeST    & 0.8590 & 0.8000 & 0.8456 & 0.4007 & 0.1832 & & 0.7261 & 0.5612 \\
\hline
Muril-LIME    & 0.8269 & 0.7679 & 0.8149 & 0.3124 & 0.1588 & & 0.5585 & 0.5517 \\
Muril-XMuTeST    & 0.8269 & 0.7679 & 0.8149 & 0.4182 & 0.1734 & & 0.5486 & 0.5615 \\
Muril-Rationale-LIME   & 0.8357 & 0.7716 & 0.8217 & 0.3160 & 0.1724 & & 0.6890 & 0.4936 \\
Muril-Rationale-XMuTeST   & 0.8357 & 0.7716 & 0.8217 & 0.4117 & 0.1906 & & 0.6945 & 0.5042 \\
\hline
GPT-4o     & 0.5461 & 0.6071 & 0.5542 & 0.3942 & 0.2266 & & - & - \\
Llama-3.1  & 0.5172 & 0.5829 & 0.5049 & 0.4263 & \textbf{0.3063} & & - & - \\
Mistral-24.07 & 0.7159 & 0.6684 & 0.7099 & 0.3975 & 0.2832 & & - & - \\
\hline
BERT-HateXplain-LIME & 0.8557 & 0.7884 & 0.8376  & 0.2822 & 0.1470 & & 0.6481 & \textbf{0.2111} \\
BERT-ILP & 0.8395 & 0.7918 & 0.8302  & 0.3941 & 0.2618 & & 0.6299 & 0.2704 \\
\midrule
\textbf{X-MuTeST with LLM} & \textbf{0.8745} & \textbf{0.8168} & \textbf{0.8623} & \textbf{0.4344} & 0.2667 & & \textbf{0.7483} & 0.4768 \\
\bottomrule
\end{tabular}
\caption{Classification and explainability performance of various methods on Hindi}
\label{table:hindi}
\end{table*}

\begin{table*}[!ht]
\centering
\scriptsize
\renewcommand{\arraystretch}{1.1}
\begin{tabular}{l ccc ccc cc}
\hline
\multirow{2}{*}{\textbf{Model}} & \multicolumn{3}{c}{\textbf{Performance}} & \multicolumn{3}{c}{\textbf{Plausibility}} & \multicolumn{2}{c}{\textbf{Faithfulness}} \\
\cmidrule(lr){2-4} \cmidrule(lr){5-7} \cmidrule(lr){8-9}
 & \textbf{Acc $\uparrow$} & \textbf{F1 $\uparrow$} & \textbf{Macro-F1 $\uparrow$} & \textbf{Token-F1 $\uparrow$} & \textbf{IOU-F1 $\uparrow$} & & \textbf{Comp $\uparrow$} & \textbf{Suff $\downarrow$} \\
 \hline
XLMR-LIME     & 0.8223 & 0.8609 & 0.8075 & 0.3886 & 0.2279 & & 0.6500 & 0.1244 \\
XLMR-XMuTeST     & 0.8223 & 0.8609 & 0.8075 & 0.4087 & 0.2195 & & 0.6842 & 0.0945 \\
XLMR-Rationale-LIME    & 0.8349 & 0.8673 & 0.8245 & 0.4665 & 0.2389 & & 0.7449 & 0.0559 \\
XLMR-Rationale-XMuTeST    & 0.8349 & 0.8673 & 0.8245 & 0.4752 & 0.2357 & & 0.7682 & 0.0546 \\ 
\hline
Muril-LIME    & 0.8307 & 0.8694 & 0.8144 & 0.3908 & 0.2002 & & 0.7365 & 0.0898 \\
Muril-XMuTeST    & 0.8307 & 0.8694 & 0.8144 & 0.3928 & 0.2184 & & 0.7527 & 0.0871 \\
Muril-Rationale-LIME   & 0.8433 & 0.8744 & 0.8385 & 0.4553 & 0.2121 & & 0.7992 & 0.0856 \\
Muril-Rationale-XMuTeST   & 0.8433 & 0.8744 & 0.8385 & 0.4743 & 0.2195 & & 0.8372 & 0.0795 \\ 
\hline
GPT-4o     & 0.7802 & 0.8266 & 0.7633 & 0.4613 & 0.3097 & & - & - \\
Llama-3.1  & 0.7560 & 0.8086 & 0.7362 & 0.3995 & 0.2819 & & - & - \\
Mistral-24.07 & 0.7918 & 0.8448 & 0.7643 & 0.3934 & \textbf{0.3176} & & - & - \\ 
\hline
BERT-HateXplain-LIME & 0.8339 & 0.8647 & 0.8247  & 0.4454 & 0.2382 & & \textbf{0.8555} & 0.2019 \\
BERT-ILP & 0.8145 & 0.7967 & 0.8028  & 0.4387 & 0.2291 & & 0.7824 & 0.2008 \\
\midrule
\textbf{X-MuTeST with LLM} & \textbf{0.8604} & \textbf{0.8827} & \textbf{0.8513} & \textbf{0.5125} & 0.3086 & & 0.8433 & \textbf{0.0507} \\
\bottomrule
\end{tabular}
\caption{Classification and explainability performance of various methods on English}
\label{table:english}
\end{table*}

\subsection{Results Analysis for Telugu}

Table~\ref{table:telugu} presents the classification and explanation performance of various models on the Telugu dataset. The LLM-consulted X-MuTeST outperforms all other models across metrics. It achieves the highest accuracy (0.8881), F1 score (0.8762), and Macro-F1 (0.8849), surpassing the next-best baseline, Muril-RX, by 1.41\% in accuracy and 1.29\% in Macro-F1. These results highlight the effectiveness of integrating LLM-consulted explanations combined with X-MuTeST.

On explainability metrics, X-MuTeST yields the best plausibility performance, with token-F1 of 0.6231 and IOU-F1 of 0.3189, improvements of 4.77\% and 2.76\%, respectively, over Muril-RX. This indicates stronger alignment of the generated rationales with human annotations.
In terms of faithfulness, X-MuTeST obtains a lower (better) sufficiency score of 0.0448 compared to 0.0747 for Muril-RX, suggesting that its identified rationales are more essential to the model's prediction.
While Muril-RX performs competitively, X-MuTeST's two-stage learning process, combining n-gram-based attention improvement with LLM-generated explanation fusion, leads to superior interpretability and confidence. In contrast, LLMs such as GPT-4o, Llama-3.1, and Mistral-24.07 show lower performance in both classification and explainability, likely due to inaccurate transliteration from Latin to Telugu script.

\begin{figure*}[h] 
    \centering
    \includegraphics[width=1.0\textwidth]{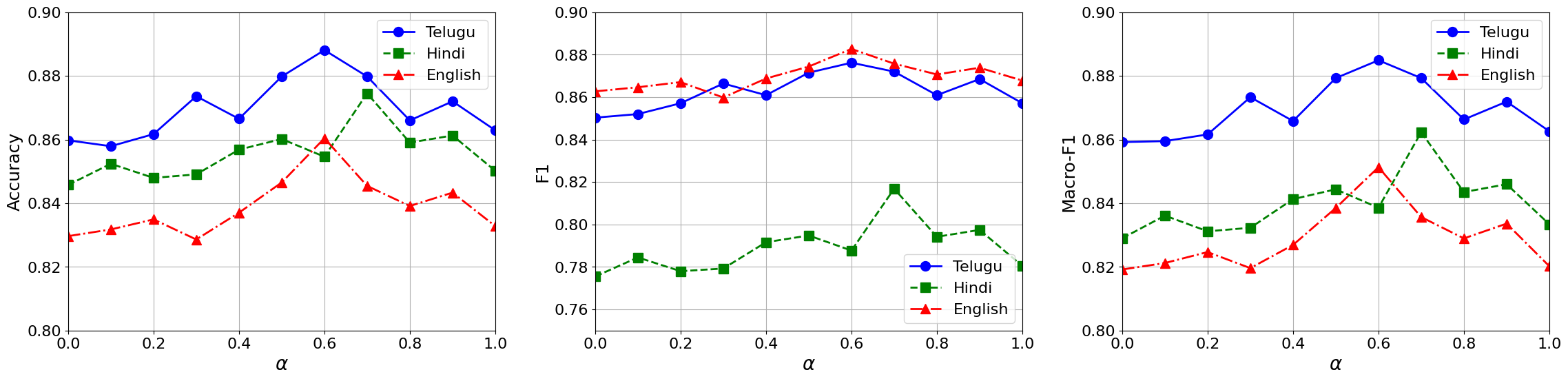} 
    \caption{Influence of $\alpha$ parameter on classification performance in stage-2 training}
    \label{fig:graph}
\end{figure*}

\begin{figure}[h] 
    \centering
    \includegraphics[width=0.35\textwidth]{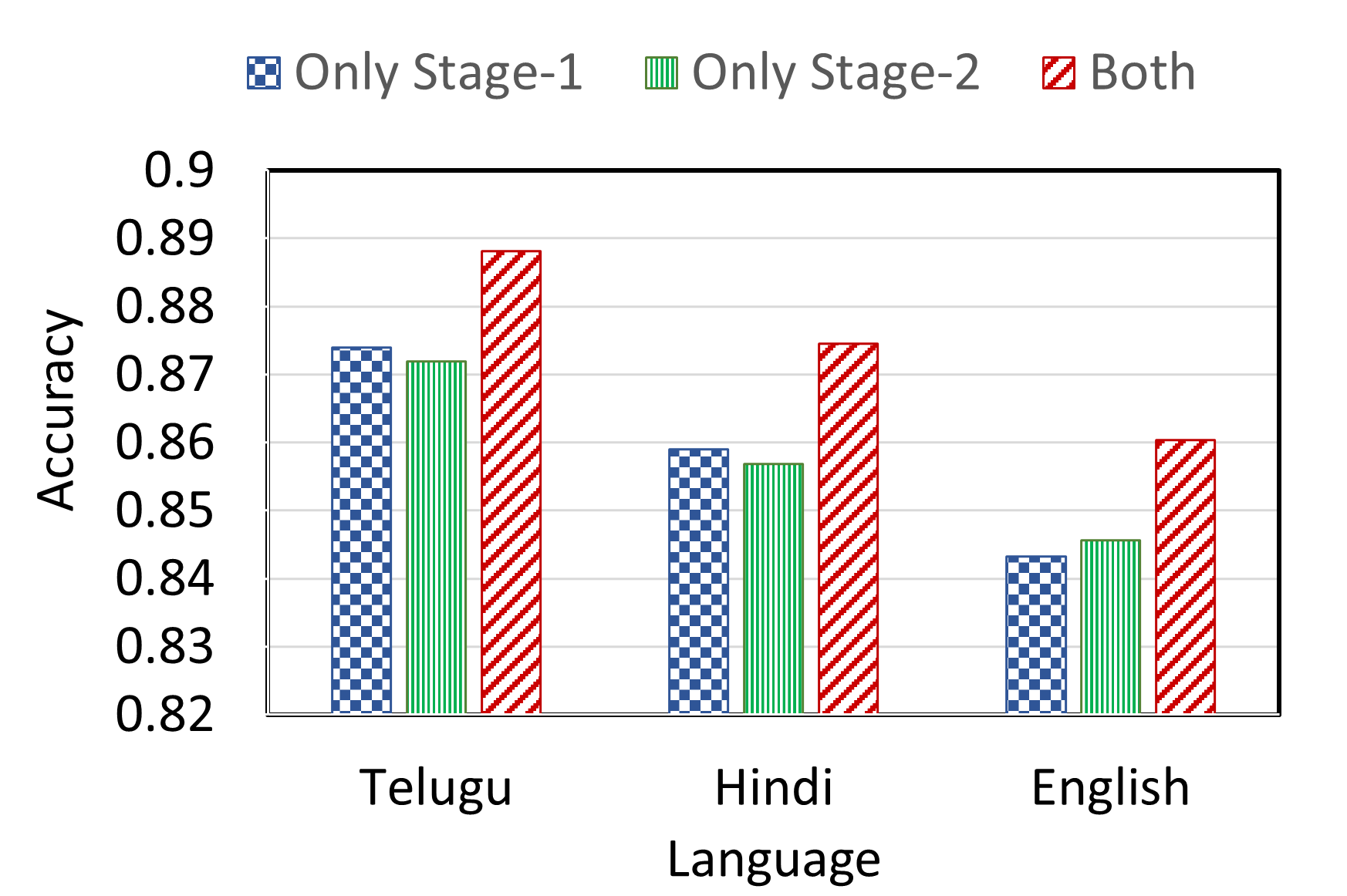} 
    \caption{Ablation on two stages of training}
    \label{fig:training}
\end{figure}

\subsection{Results Analysis for Hindi}
\label{sec:hindi_results}

Table~\ref{table:hindi} shows that X-MuTeST with LLM achieves the highest classification and explanation scores for Hindi, with an accuracy of 0.8745 and token-F1 of 0.4344. Compared to the next-best model (Muril-RX), it gains 3.88\% in accuracy and 2.27\% in token-F1. It also improves sufficiency (0.4768 vs. 0.5042), showing more faithful explanations.
While LLMs like Llama-3.1 show high token-F1 (0.4263), their classification drops sharply by 35.73\%, indicating inaccurate token lists. XLMR and BERT-based baselines also trail across all metrics. These results confirm that the LLM-guided two-stage training in X-MuTeST produces more accurate and human-aligned explanations.

\subsection{Results Analysis for English}
\label{sec:eng_results}

Consistent with the findings for Telugu and Hindi, X-MuTeST achieves the highest classification performance in English as shown in Table \ref{table:english}. In terms of explainability, X-MuTeST leads with a token-F1 score of 0.5125 and demonstrates strong faithfulness with a high comprehensiveness of 0.8433 and the lowest sufficiency score of 0.0507.
LLMs show moderate classification performance, with GPT-4o scoring relatively high in token-F1 of 0.4613 but lower in overall classification accuracy (0.7802). Mistral-24.07
achieves the highest IOU-F1 of 0.3176, though its overall performance lags behind X-MuTeST.
Overall, X-MuTeST consistently outperforms other models in classification while maintaining competitive scores in explainability for the English dataset, too.

\subsection{Parameter Sensitivity Analysis \& Ablation}
\label{sec:alpha}

The graphs in Figure \ref{fig:graph} show the influence of $\alpha$ on the classification performance in different languages. The $\alpha$ parameter denotes the weights given to attention loss, whereas (1-$\alpha$) weights are given to classification loss. The attention loss is computed between the attention mask generated through the X-MuTeST explainability method and the predicted average attention score for each token relative to the \([CLS]\) token. For Hindi and English, the best results are achieved when $\alpha$ is set to 0.6, denoting that the X-MuTeST’s attention mask aligns well with that of the model. Lower performance is noted on the boundary conditions when either the entire weight is given to attention loss or the classification loss, denoting that integration of the X-MuTeST explainability method strikes a balance, resulting in enhanced performance. The best performance for Telugu is noted when $\alpha$ is set to 0.7. 
Additionally, Figure \ref{fig:training} shows the ablation study results on two stages of training. Performance degradation is observed on the removal of any of the two stages of training as depicted in the graph. This further demonstrates that the proposed training method of X-MuTeST enhances the classification performance while improving the explainability.

\subsection{Generalizability}

To assess the explanation generalizability of the proposed framework, we conducted additional experiments on two benchmark datasets: HateXplain~\cite{mathew2021hatexplain} and HateBRXplain~\cite{salles2025hatebrxplain}. On the HateXplain dataset, X-MuTeST achieved an IOU-F1 of 0.314 and a Token-F1 of 0.587, surpassing the best reported results in this work by 9.2\% and 8.1\%, respectively. It also yielded a 5.6\% improvement in comprehensiveness. Similarly, on the HateBRXplain dataset, X-MuTeST with LLM recorded a 6.9\% gain in Token-F1 and a 2.3\% improvement in sufficiency over the strongest baseline. These results demonstrate that the proposed framework generalizes effectively across diverse datasets and settings.

\section{Conclusion}
\label{sec:conclusion}
In this study, we present rationale resources and the LLM-consulted X-MuTeST framework, which demonstrates robust classification and explainability performance across three languages, Hindi, Telugu, and English, for the task of hate speech detection. By incorporating a combination of traditional classification metrics and explainability measures, the framework not only improved classification accuracy but also provided deeper insights into model behaviour. The results suggest that two-stage training and integration of explainability into multilingual text classification enhances both model trustworthiness and user understanding. Further investigations could involve expanding the framework to handle more languages.

\section{Acknowledgments}
 We used two datasets, HASOC (2020 and 2021) and HOLD. We would like to acknowledge the contributions of the HASOC and HOLD teams. These datasets were instrumental in conducting this research.

\bibliography{aaai2026}

\begin{thebibliography}{41}
\providecommand{\natexlab}[1]{#1}

\bibitem[{Arshad and Shahzad(2024)}]{arshad2024understanding}
Arshad, M.~U.; and Shahzad, W. 2024.
\newblock Understanding hate speech: the HateInsights dataset and model interpretability.
\newblock \emph{PeerJ Computer Science}, 10: e2372.

\bibitem[{Camburu et~al.(2018)Camburu, Rockt{\"a}schel, Lukasiewicz, and Blunsom}]{camburu2018snli}
Camburu, O.-M.; Rockt{\"a}schel, T.; Lukasiewicz, T.; and Blunsom, P. 2018.
\newblock e-snli: Natural language inference with natural language explanations.
\newblock \emph{Advances in Neural Information Processing Systems}, 31.

\bibitem[{Chakravarthi et~al.(2024)Chakravarthi, Priyadharshini, Thavareesan, Sherly, Nadarajan, Ravikiran et~al.}]{chakravarthi2024proceedings}
Chakravarthi, B.~R.; Priyadharshini, R.; Thavareesan, S.; Sherly, E.; Nadarajan, R.; Ravikiran, M.; et~al. 2024.
\newblock Proceedings of the Fourth Workshop on Speech, Vision, and Language Technologies for Dravidian Languages.
\newblock In \emph{Proceedings of the Fourth Workshop on Speech, Vision, and Language Technologies for Dravidian Languages}.

\bibitem[{Clark et~al.(2019)Clark, Lee, Chang, Kwiatkowski, Collins, and Toutanova}]{clark2019boolq}
Clark, C.; Lee, K.; Chang, M.-W.; Kwiatkowski, T.; Collins, M.; and Toutanova, K. 2019.
\newblock BoolQ: Exploring the Surprising Difficulty of Natural Yes/No Questions.
\newblock In \emph{Proceedings of NAACL-HLT}, 2924--2936.

\bibitem[{Clarke et~al.(2023)Clarke, Hall, Mittal, Yu, Sajeev, Mars, and Chen}]{clarke2023rule}
Clarke, C.; Hall, M.; Mittal, G.; Yu, Y.; Sajeev, S.; Mars, J.; and Chen, M. 2023.
\newblock Rule By Example: Harnessing Logical Rules for Explainable Hate Speech Detection.
\newblock In \emph{Proceedings of the 61st Annual Meeting of the Association for Computational Linguistics (Volume 1: Long Papers)}, 364--376.

\bibitem[{Conneau et~al.(2020)Conneau, Khandelwal, Goyal, Chaudhary, Wenzek, Guzm{\'a}n, Grave, Ott, Zettlemoyer, and Stoyanov}]{conneau2020unsupervised}
Conneau, A.; Khandelwal, K.; Goyal, N.; Chaudhary, V.; Wenzek, G.; Guzm{\'a}n, F.; Grave, E.; Ott, M.; Zettlemoyer, L.; and Stoyanov, V. 2020.
\newblock Unsupervised Cross-lingual Representation Learning at Scale.
\newblock In Jurafsky, D.; Chai, J.; Schluter, N.; and Tetreault, J., eds., \emph{Proceedings of the 58th Annual Meeting of the Association for Computational Linguistics}, 8440--8451. Online: Association for Computational Linguistics.

\bibitem[{Devlin et~al.(2019)Devlin, Chang, Lee, and Toutanova}]{devlin2019bert}
Devlin, J.; Chang, M.-W.; Lee, K.; and Toutanova, K. 2019.
\newblock {BERT}: Pre-training of Deep Bidirectional Transformers for Language Understanding.
\newblock In Burstein, J.; Doran, C.; and Solorio, T., eds., \emph{Proceedings of the 2019 Conference of the North {A}merican Chapter of the Association for Computational Linguistics: Human Language Technologies, Volume 1 (Long and Short Papers)}, 4171--4186. Minneapolis, Minnesota: Association for Computational Linguistics.

\bibitem[{DeYoung et~al.(2020)DeYoung, Jain, Rajani, Lehman, Xiong, Socher, and Wallace}]{deyoung2020eraser}
DeYoung, J.; Jain, S.; Rajani, N.~F.; Lehman, E.; Xiong, C.; Socher, R.; and Wallace, B.~C. 2020.
\newblock ERASER: A Benchmark to Evaluate Rationalized NLP Models.
\newblock In \emph{Proceedings of the 58th Annual Meeting of the Association for Computational Linguistics}, 4443--4458.

\bibitem[{Dubey et~al.(2024)Dubey, Jauhri, Pandey, Kadian, Al-Dahle, Letman, Mathur, Schelten, Yang, Fan et~al.}]{dubey2024llama}
Dubey, A.; Jauhri, A.; Pandey, A.; Kadian, A.; Al-Dahle, A.; Letman, A.; Mathur, A.; Schelten, A.; Yang, A.; Fan, A.; et~al. 2024.
\newblock The llama 3 herd of models.
\newblock \emph{arXiv preprint arXiv:2407.21783}.

\bibitem[{Geleta et~al.(2023)Geleta, Eckelt, Parada-Cabaleiro, and Schedl}]{geleta2023exploring}
Geleta, R.~R.; Eckelt, K.; Parada-Cabaleiro, E.; and Schedl, M. 2023.
\newblock Exploring intensities of hate speech on social media: A case study on explaining multilingual models with XAI.
\newblock In \emph{Proceedings of the 4th Conference on Language, Data and Knowledge}, 532--537.

\bibitem[{Hahn et~al.(2012)Hahn, Beisswanger, Buyko, Faessler, Traum{\"u}ller, Schr{\"o}der, and Hornbostel}]{hahn2012iterative}
Hahn, U.; Beisswanger, E.; Buyko, E.; Faessler, E.; Traum{\"u}ller, J.; Schr{\"o}der, S.; and Hornbostel, K. 2012.
\newblock Iterative Refinement and Quality Checking of Annotation Guidelines—How to Deal Effectively with Semantically Sloppy Named Entity Types, such as Pathological Phenomena.
\newblock In \emph{Proceedings of the Eighth International Conference on Language Resources and Evaluation (LREC'12)}, 3881--3885.

\bibitem[{Jacovi and Goldberg(2020)}]{jacovi2020towards}
Jacovi, A.; and Goldberg, Y. 2020.
\newblock Towards Faithfully Interpretable NLP Systems: How Should We Define and Evaluate Faithfulness?
\newblock In \emph{Proceedings of the 58th Annual Meeting of the Association for Computational Linguistics}, 4198--4205.

\bibitem[{Kapil et~al.(2023)Kapil, Kumari, Ekbal, Pal, Chatterjee, and Vinutha}]{kapil2023hhld}
Kapil, P.; Kumari, G.; Ekbal, A.; Pal, S.; Chatterjee, A.; and Vinutha, B. 2023.
\newblock HHLD: Hateful posts Identification in Hindi Language leveraging multi task learning.
\newblock \emph{IEEE Access}.

\bibitem[{Kasu et~al.(2025)Kasu, Rehman, Dar, Bharat~Junghare, Namboodiri, and Kumar}]{kasu2025d}
Kasu, S. K.~R.; Rehman, M. Z.~U.; Dar, S.~S.; Bharat~Junghare, R.; Namboodiri, D.~S.; and Kumar, N. 2025.
\newblock D-humor: Dark humor understanding via multimodal open-ended reasoning.
\newblock \emph{arXiv e-prints}, arXiv--2509.

\bibitem[{Khanuja et~al.(2021)Khanuja, Bansal, Mehtani, Khosla, Dey, Gopalan, Margam, Aggarwal, Nagipogu, Dave et~al.}]{khanuja2021muril}
Khanuja, S.; Bansal, D.; Mehtani, S.; Khosla, S.; Dey, A.; Gopalan, B.; Margam, D.~K.; Aggarwal, P.; Nagipogu, R.~T.; Dave, S.; et~al. 2021.
\newblock Muril: Multilingual representations for indian languages.
\newblock \emph{arXiv preprint arXiv:2103.10730}.

\bibitem[{Khashabi et~al.(2018)Khashabi, Chaturvedi, Roth, Upadhyay, and Roth}]{khashabi2018looking}
Khashabi, D.; Chaturvedi, S.; Roth, M.; Upadhyay, S.; and Roth, D. 2018.
\newblock Looking beyond the surface: A challenge set for reading comprehension over multiple sentences.
\newblock In \emph{Proceedings of the 2018 Conference of the North American Chapter of the Association for Computational Linguistics: Human Language Technologies, Volume 1 (Long Papers)}, 252--262.

\bibitem[{Lehman et~al.(2019)Lehman, DeYoung, Barzilay, and Wallace}]{lehman2019inferring}
Lehman, E.; DeYoung, J.; Barzilay, R.; and Wallace, B.~C. 2019.
\newblock Inferring Which Medical Treatments Work from Reports of Clinical Trials.
\newblock In \emph{Proceedings of the 2019 Conference of the North American Chapter of the Association for Computational Linguistics: Human Language Technologies, Volume 1 (Long and Short Papers)}, 3705--3717.

\bibitem[{Lin et~al.(2024)Lin, Luo, Gao, Ma, Wang, and Yang}]{lin2024towards}
Lin, H.; Luo, Z.; Gao, W.; Ma, J.; Wang, B.; and Yang, R. 2024.
\newblock Towards explainable harmful meme detection through multimodal debate between large language models.
\newblock In \emph{Proceedings of the ACM on Web Conference 2024}, 2359--2370.

\bibitem[{Madhani et~al.(2023)Madhani, Parthan, Bedekar, Nc, Khapra, Kunchukuttan, Kumar, and Khapra}]{madhani2023aksharantar}
Madhani, Y.; Parthan, S.; Bedekar, P.; Nc, G.; Khapra, R.; Kunchukuttan, A.; Kumar, P.; and Khapra, M.~M. 2023.
\newblock Aksharantar: Open Indic-language transliteration datasets and models for the next billion users.
\newblock In \emph{Findings of the Association for Computational Linguistics: EMNLP 2023}, 40--57.

\bibitem[{Mandl et~al.(2020{\natexlab{a}})Mandl, Modha, Kumar~M, and Chakravarthi}]{mandl2020overview}
Mandl, T.; Modha, S.; Kumar~M, A.; and Chakravarthi, B.~R. 2020{\natexlab{a}}.
\newblock Overview of the hasoc track at fire 2020: Hate speech and offensive language identification in tamil, malayalam, hindi, english and german.
\newblock In \emph{Proceedings of the 12th annual meeting of the forum for information retrieval evaluation}, 29--32.

\bibitem[{Mandl et~al.(2020{\natexlab{b}})Mandl, Modha, Shahi, Jaiswal, Nandini, Patel, Majumder, and Sch{\"a}fer}]{mandl2020overview1}
Mandl, T.; Modha, S.; Shahi, G.~K.; Jaiswal, A.~K.; Nandini, D.; Patel, D.; Majumder, P.; and Sch{\"a}fer, J. 2020{\natexlab{b}}.
\newblock Overview of the HASOC track at FIRE 2020: Hate Speech and Offensive Content Identification in Indo-European Languages.

\bibitem[{Mathew et~al.(2021)Mathew, Saha, Yimam, Biemann, Goyal, and Mukherjee}]{mathew2021hatexplain}
Mathew, B.; Saha, P.; Yimam, S.~M.; Biemann, C.; Goyal, P.; and Mukherjee, A. 2021.
\newblock Hatexplain: A benchmark dataset for explainable hate speech detection.
\newblock In \emph{Proceedings of the AAAI conference on artificial intelligence}, volume~35, 14867--14875.

\bibitem[{MistralAI(2024)}]{mistral}
MistralAI. 2024.
\newblock {L}arge {E}nough --- mistral.ai.
\newblock \url{https://mistral.ai/news/mistral-large-2407/}.

\bibitem[{Modha et~al.(2021)Modha, Mandl, Shahi, Madhu, Satapara, Ranasinghe, and Zampieri}]{modha2021overview}
Modha, S.; Mandl, T.; Shahi, G.~K.; Madhu, H.; Satapara, S.; Ranasinghe, T.; and Zampieri, M. 2021.
\newblock Overview of the hasoc subtrack at fire 2021: Hate speech and offensive content identification in english and indo-aryan languages and conversational hate speech.
\newblock In \emph{Proceedings of the 13th Annual Meeting of the Forum for Information Retrieval Evaluation}, 1--3.

\bibitem[{Nguyen and Rudra(2022)}]{nguyen2022towards}
Nguyen, T.~H.; and Rudra, K. 2022.
\newblock Towards an interpretable approach to classify and summarize crisis events from microblogs.
\newblock In \emph{Proceedings of the ACM Web Conference 2022}, 3641--3650.

\bibitem[{OpenAI(2024)}]{gpt}
OpenAI. 2024.
\newblock {C}hat{G}{P}{T} 4o.
\newblock url : \url{https://platform.openai.com/docs/models/gpt-4o}.

\bibitem[{Premjith et~al.(2024)Premjith, Chakravarthi, Kumaresan, Rajiakodi, Karnati, Mangamuru, and Janakiram}]{premjith2024findings}
Premjith, B.; Chakravarthi, B.~R.; Kumaresan, P.~K.; Rajiakodi, S.; Karnati, S.; Mangamuru, S.; and Janakiram, C. 2024.
\newblock Findings of the Shared Task on Hate and Offensive Language Detection in Telugu Codemixed Text (HOLD-Telugu)@ DravidianLangTech 2024.
\newblock In \emph{Proceedings of the Fourth Workshop on Speech, Vision, and Language Technologies for Dravidian Languages}, 49--55.

\bibitem[{Priyadharshini et~al.(2023)Priyadharshini, Chakravarthi, Malliga, Subalalitha, Kogilavani, Premjith, Murugappan, and Kumaresan}]{priyadharshini2023overview}
Priyadharshini, R.; Chakravarthi, B.~R.; Malliga, S.; Subalalitha, C.; Kogilavani, S.; Premjith, B.; Murugappan, A.; and Kumaresan, P.~K. 2023.
\newblock Overview of Shared-task on Abusive Comment Detection in Tamil and Telugu.
\newblock In \emph{Proceedings of the Third Workshop on Speech and Language Technologies for Dravidian Languages}, 80--87.

\bibitem[{Rajani et~al.(2019)Rajani, McCann, Xiong, and Socher}]{rajani2019explain}
Rajani, N.~F.; McCann, B.; Xiong, C.; and Socher, R. 2019.
\newblock Explain Yourself! Leveraging Language Models for Commonsense Reasoning.
\newblock In \emph{Proceedings of the 57th Annual Meeting of the Association for Computational Linguistics}, 4932--4942.

\bibitem[{Rehman et~al.(2025)Rehman, Bhatnagar, Kabde, Bansal, and Kumar}]{rehman2025implihatevid}
Rehman, M. Z.~U.; Bhatnagar, A.; Kabde, O.; Bansal, S.; and Kumar, N. 2025.
\newblock ImpliHateVid: A Benchmark Dataset and Two-stage Contrastive Learning Framework for Implicit Hate Speech Detection in Videos.
\newblock In \emph{Proceedings of the 63rd Annual Meeting of the Association for Computational Linguistics (Volume 1: Long Papers)}, 17209--17221.

\bibitem[{Rehman et~al.(2023)Rehman, Mehta, Singh, Kaushik, and Kumar}]{rehman2023user}
Rehman, M. Z.~U.; Mehta, S.; Singh, K.; Kaushik, K.; and Kumar, N. 2023.
\newblock User-aware multilingual abusive content detection in social media.
\newblock \emph{Information Processing \& Management}, 60(5): 103450.

\bibitem[{Roy et~al.(2023)Roy, Harshvardhan, Mukherjee, and Saha}]{roy2023probing}
Roy, S.; Harshvardhan, A.; Mukherjee, A.; and Saha, P. 2023.
\newblock Probing LLMs for hate speech detection: strengths and vulnerabilities.
\newblock In \emph{Findings of the Association for Computational Linguistics: EMNLP 2023}, 6116--6128.

\bibitem[{Salles, Vargas, and Benevenuto(2025)}]{salles2025hatebrxplain}
Salles, I.; Vargas, F.; and Benevenuto, F. 2025.
\newblock HateBRXplain: A Benchmark Dataset with Human-Annotated Rationales for Explainable Hate Speech Detection in Brazilian Portuguese.
\newblock In \emph{Proceedings of the 31st International Conference on Computational Linguistics}, 6659--6669.

\bibitem[{Sawant et~al.(2024)Sawant, Younus, Caton, and Qureshi}]{sawant2024using}
Sawant, M.; Younus, A.; Caton, S.; and Qureshi, M.~A. 2024.
\newblock Using Explainable AI (XAI) for Identification of Subjectivity in Hate Speech Annotations for Low-Resource Languages.
\newblock In \emph{Proceedings of the 4th International Workshop on Open Challenges in Online Social Networks}, 10--17.

\bibitem[{Schmer-Galunder et~al.(2024)Schmer-Galunder, Wheelock, Jalan, Chvasta, Friedman, and Saltz}]{schmer2024annotator}
Schmer-Galunder, S.; Wheelock, R.; Jalan, Z.; Chvasta, A.; Friedman, S.; and Saltz, E. 2024.
\newblock Annotator in the Loop: A Case Study of In-Depth Rater Engagement to Create a Prosocial Benchmark Dataset.
\newblock In \emph{Proceedings of the AAAI/ACM Conference on AI, Ethics, and Society}, volume~7, 1319--1328.

\bibitem[{Wang et~al.(2024)Wang, Ma, Lin, Yang, Yang, Tian, and Chang}]{wang2024explainable}
Wang, B.; Ma, J.; Lin, H.; Yang, Z.; Yang, R.; Tian, Y.; and Chang, Y. 2024.
\newblock Explainable Fake News Detection With Large Language Model via Defense Among Competing Wisdom.
\newblock In \emph{Proceedings of the ACM on Web Conference 2024}, 2452--2463.

\bibitem[{Waseem and Hovy(2016)}]{waseem2016hateful}
Waseem, Z.; and Hovy, D. 2016.
\newblock Hateful symbols or hateful people? predictive features for hate speech detection on twitter.
\newblock In \emph{Proceedings of the NAACL student research workshop}, 88--93.

\bibitem[{Yadav, Kaushik, and McDaid(2023)}]{yadav2023hate}
Yadav, S.; Kaushik, A.; and McDaid, K. 2023.
\newblock Hate speech is not free speech: Explainable machine learning for hate speech detection in code-mixed languages.
\newblock In \emph{2023 IEEE International Symposium on Technology and Society (ISTAS)}, 1--8. IEEE.

\bibitem[{Yessenalina, Choi, and Cardie(2010)}]{yessenalina2010automatically}
Yessenalina, A.; Choi, Y.; and Cardie, C. 2010.
\newblock Automatically generating annotator rationales to improve sentiment classification.
\newblock In \emph{Proceedings of the ACL 2010 Conference Short Papers}, 336--341.

\bibitem[{Zaidan and Eisner(2008)}]{zaidan2008modeling}
Zaidan, O.; and Eisner, J. 2008.
\newblock Modeling annotators: A generative approach to learning from annotator rationales.
\newblock In \emph{Proceedings of the 2008 conference on Empirical methods in natural language processing}, 31--40.

\bibitem[{Zaidan, Eisner, and Piatko(2007)}]{zaidan2007using}
Zaidan, O.; Eisner, J.; and Piatko, C. 2007.
\newblock Using “annotator rationales” to improve machine learning for text categorization.
\newblock In \emph{Human language technologies 2007: The conference of the North American chapter of the association for computational linguistics; proceedings of the main conference}, 260--267.

\end{thebibliography}

\end{document}